\documentclass[runningheads, envcountsame, a4paper]{llncs}
\usepackage{graphicx}
\usepackage{subcaption}
\usepackage{tabularx,booktabs}
\usepackage{makecell}
\usepackage{todonotes}
\usepackage{amssymb}
\usepackage{amsmath}
\usepackage[misc]{ifsym}
\newcolumntype{Y}{>{\centering\arraybackslash}X}
\newcommand{\printfnsymbol}[1]{%
  \textsuperscript{\@fnsymbol{#1}}%
}
\usepackage{hyperref}

\begin{document}
\title{An Uncertainty-based Human-in-the-loop System for Industrial Tool Wear Analysis}
\toctitle{An uncertainty-based human-in-the-loop system for industrial tool wear analysis}
\titlerunning{Uncertainty-based Human-in-the-loop System}
%

\author{
Alexander Treiss\thanks{Alexander Treiss and Jannis Walk contributed equally in a shared first authorship.}\Letter \and Jannis Walk\textsuperscript{$\star$}\Letter 
\and Niklas K\"uhl}
\tocauthor{Alexander~Treiss, Jannis~Walk and Niklas~K\"uhl}
\authorrunning{A. Treiss, J. Walk, and N. K\"uhl}
%
\institute{Karlsruhe Institute of Technology, Kaiserstrasse 89, 76133 Karlsruhe, Germany \\ \email{alexander.treiss@alumni.kit.edu, \{walk, kuehl\}@kit.edu}}
\maketitle              

\begin{abstract}
Convolutional neural networks have shown to achieve superior performance on image segmentation tasks. However, convolutional neural networks, operating as black-box systems, generally do not provide a reliable measure about the confidence of their decisions. This leads to various problems in industrial settings, amongst others, inadequate levels of trust from users in the model's outputs as well as a non-compliance with current policy guidelines (e.g., EU AI Strategy). To address these issues, we use uncertainty measures based on Monte-Carlo dropout in the context of a human-in-the-loop system to increase the system's transparency and performance.  In particular, we demonstrate the benefits described above on a real-world multi-class image segmentation task of wear analysis in the machining industry. Following previous work, we show that the quality of a prediction correlates with the model's uncertainty. Additionally, we demonstrate that a multiple linear regression using the model's uncertainties as independent variables significantly explains the quality of a prediction (\(R^2=0.718\)). Within the uncertainty-based human-in-the-loop system, the multiple regression aims at identifying failed predictions on an image-level. The system utilizes a human expert to label these failed predictions manually. A simulation study demonstrates that the uncertainty-based human-in-the-loop system increases performance for different levels of human involvement in comparison to a random-based human-in-the-loop system. To ensure generalizability, we show that the presented approach achieves similar results on the publicly available Cityscapes dataset.


\end{abstract}

\section{Introduction}
\begin{figure}[ht] 
    \centering
   \begin{subfigure}[b]{0.25\linewidth}
    \centering
    \includegraphics[width=1\linewidth]{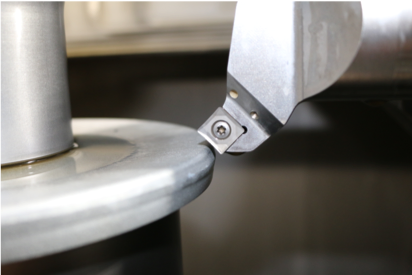} 
    \caption{Process} 
    \label{machining}
  \end{subfigure}
    \hfill
  \begin{subfigure}[b]{0.22\linewidth}
    \centering
    \includegraphics[width=1\linewidth]{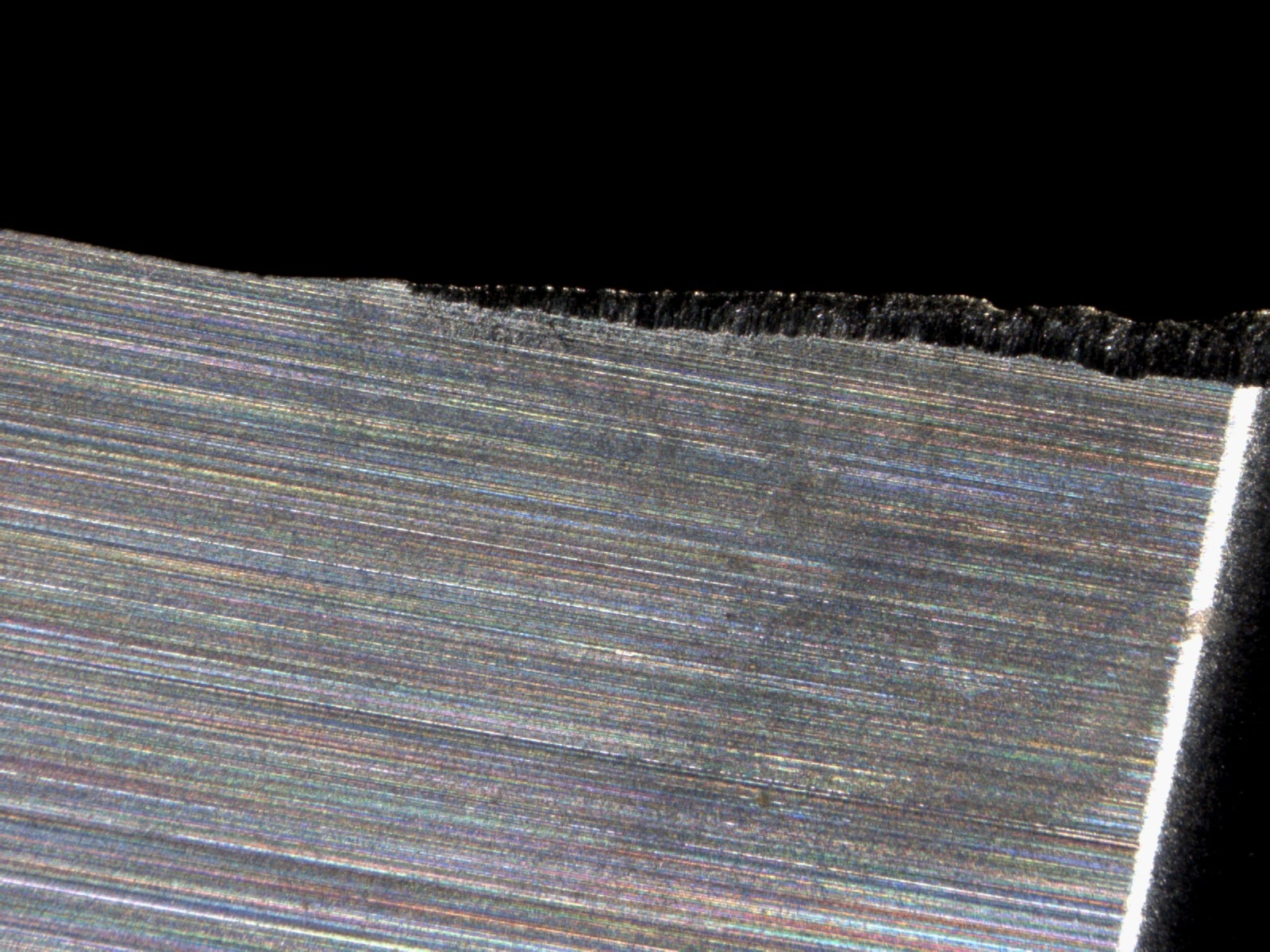} 
    \caption{Flank wear} 
    \label{flank_wear} 
  \end{subfigure}
  \hfill
  \begin{subfigure}[b]{0.22\linewidth}
    \centering
    \includegraphics[width=1\linewidth]{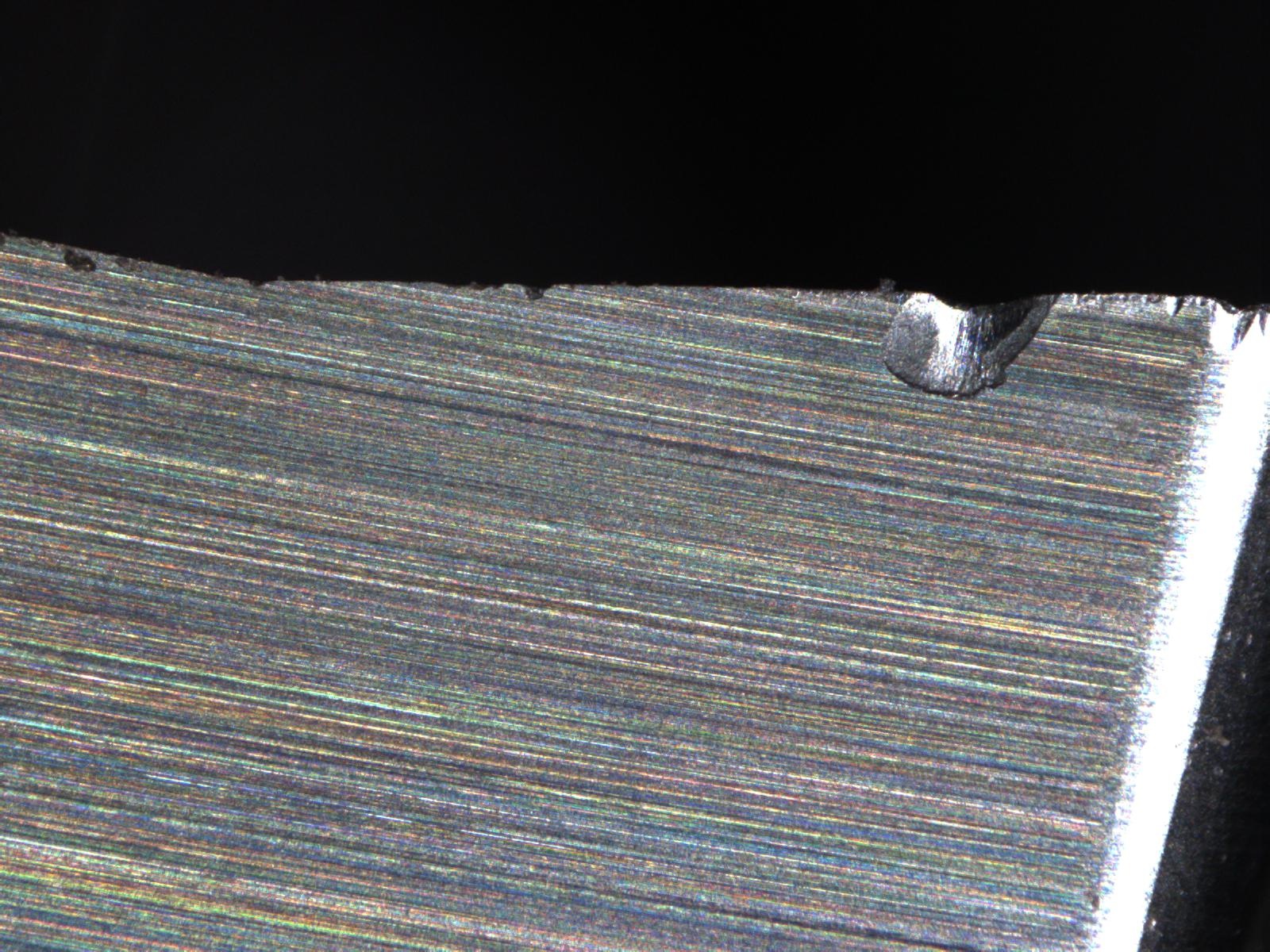} 
    \caption{Chipping} 
    \label{chipping} 
  \end{subfigure} 
  \hfill
  \begin{subfigure}[b]{0.22\linewidth}
    \centering
    \includegraphics[width=1\linewidth]{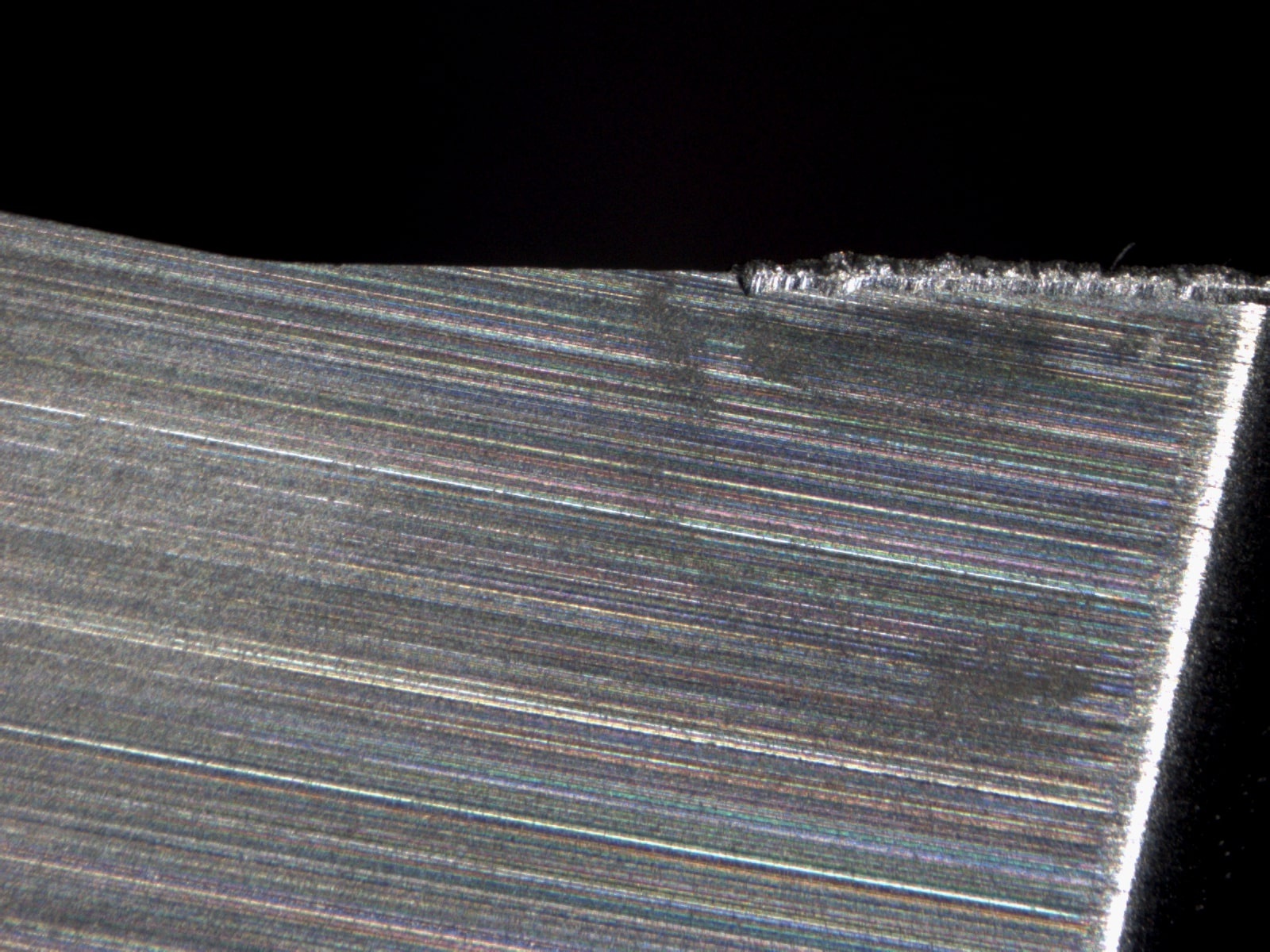} 
    \caption{Built-up edge} 
    \label{built-up-edge}
  \end{subfigure}
    \hfill
  \caption{Machining process and common wear mechanisms}
  \label{fig:examined_wear} 
\end{figure}

Machining is an essential manufacturing process \cite{Altintas2012}, which is applied in many industries, such as aerospace, automotive, and the energy and electronics industry. In general, machining describes the process of removing unwanted material from a workpiece \cite{black1995introduction}. Thereby, a cutting tool is moved in a relative motion to the workpiece to produce the desired shape  \cite{boothroyd1988fundamentals}. Figure \ref{machining} displays an exemplary image of a machining process applying a \textit{cutting tool insert}.
Cutting tools are consumables because of the occurrence of wear on the tools, which ultimately results in unusable tools.
In the following, we will briefly describe three common wear mechanisms, compare Figures \ref{flank_wear} -- \ref{built-up-edge} for exemplary images.
\textit{Flank wear} occurs due to friction between the tools flank surface and the workpiece \cite{Altintas2012}. It is unavoidable and thus the most commonly observed wear mechanism \cite{Siddhpura2013}. \textit{Chipping} refers to a set of particles breaking off from the tool's cutting edge \cite{Altintas2012}.
A \textit{built-up edge} arises when workpiece material deposits on the cutting edge due to localized high temperatures and extreme pressures \cite{black1995introduction}. 
Chipping and built-up edge are less desirable than flank wear since they induce a more severe and sudden deformation of the tool's cutting edge, leading to a reduced surface quality on the workpiece. Ultimately, this can lead to an increase of scrap components.
A visual inspection of cutting tools enables an analysis of different wear mechanisms and provides insights into the usage behavior of cutting tools. Tool manufacturers, as well as tool end-users, can later leverage these insights to optimize the utilization of tools and to identify promising directions for the development of the next tool generations. Furthermore, an automated visual inspection enables the application of tool condition monitoring within manufacturing processes \cite{dutta2013application}. These analytics-based services possess a high economic value. Research suggests that tool failures are responsible for 20\% of production downtime in machining processes \cite{kurada1997review}. Furthermore, cutting tools and their replacement account for 3--12\% of total production cost \cite{castejon2007line}.
Due to the relevance of these analytics-based sercives, our industry partner Ceratizit Austria GmbH, a tool manufacturer, agreed to closely collaborate within the research and implementation of an automated visual inspection for tool wear analysis.

Numerous studies have examined wear analysis in the machining industry to address the need for automated wear analysis. However, the majority of the research to date focuses on traditional computer vision techniques \cite{Walk2020}. Since traditional computer vision approaches require the user to fine-tune a multitude of parameters \cite{OMahony2019}, scalability often becomes an issue. Contrary to traditional computer vision approaches, deep learning-based approaches learn the required features themselves and can, therefore, be applied to different wear problems more efficiently. Another critical advantage of deep neural networks (DNNs) is their performance. In particular, the exploitation of convolutional neural networks (CNNs) has contributed to a performance increase in several computer vision tasks, e.g., classification and segmentation. CNNs are even able to surpass human-level performance in some of these settings \cite{he2015delving}, which demonstrates that a variety of visual tasks, previously performed by humans, can be automated using CNNs. Recently, Lutz et al. (2019) \cite{Lutz2019} published the first work utilizing a CNN for wear analysis on cutting tool inserts, reporting promising results.
However, CNNs, functioning as black-box systems, generally do not provide a reliable measure about the confidence of their decisions. This shortcoming is critical because the trustworthiness of a model's output remains unclear for a human supervisor. Two scenarios \cite{lee2004trust} can unfold: First, the model's capabilities can be underestimated, resulting in a disuse of the system. Secondly, the human supervisor can overestimate the model's capabilities, leading to a misuse. The correct balance between trust and distrust constitutes one of the main barriers for a successful adoption of CNNs in many real-world use cases \cite{dellermann2019hybrid}.
In particular, a measure of confidence is essential in safety-critical applications and in scenarios for which data is limited \cite{kendall2017uncertainties}.
Limited amounts of data often occur in industrial problems, where  resources and knowledge required to label and retrieve data are frequently restricted. In these settings, CNNs can occasionally produce sub-optimal results because they usually require a substantial amount of training data. 
Moreover, while performance can be high on average, within safety-critical applications, it is crucial to filter out any erroneous output. Lastly, due to the black-box property, CNNs are currently non-compliant with the Ethics Guidelines for Trustworthy Artificial Intelligence by the European Union \cite{AI2019}. In the future, these guidelines could translate into legislation that would limit the application of CNN-based systems in some industrial settings.  

In this work, we address the need for CNN-based systems to output a confidence measure in an industrial environment. We consider the task of tool wear analysis using a unique, real-world data set from our industry partner Ceratizit. We employ an image segmentation algorithm based on the U-Net \cite{ronneberger2015u} for the pixel-wise classification of three different wear mechanisms on cutting tool inserts. To increase transparency and performance, we further enhance the tool wear analysis system with capabilities to function as an uncertainty-based human-in-the-loop system. The suggested system aims at classifying the quality of a prediction, enabling the incorporation of a human expert. In particular, we estimate the quality of a prediction using the model's uncertainty. As a foundation for uncertainty, we apply Monte-Carlo dropout (MC-dropout), which approximates a Bayesian Neural Network (BNN) \cite{gal2015dropout}. The approximated BNN outputs a probability distribution over the outputs. Based on the probability distribution, we apply multiple measures that aim at capturing the uncertainty of an output.
Subsequently, we show that for the use case of wear analysis, there exists a significant linear correlation between the uncertainty measures and the performance metric, the Dice coefficient. The linear relationship enables the utilization of the uncertainty measures as explanatory variables to predict the quality of a prediction in the absence of ground truth. We utilize these quality estimations in the following way: Predictions, which are estimated to be of high quality, are marked as successful by the system. These predictions are then passed on for automated analysis without any further human involvement. Otherwise, if an output  is marked as failed, a human expert is requested to annotate the image manually. Hence, the system is introducing transparency by measuring the confidence of the predictions and is furthermore increasing performance with the selective use of a human expert. \newline

Overall, we see the following contributions of our work. While research is carried out within the field of medical imaging, no previous study, to the best of our knowledge, has investigated the use of uncertainty estimates in order to predict segmentation quality in an industrial setting. We contribute by showing how an uncertainty-based assessment of segmentation quality can be utilized in an industrial task of tool wear analysis. While Lutz et al. (2019) \cite{Lutz2019} implement a CNN for tool wear analysis, we are additionally able to generate and leverage confidence estimates of the predictions.
Besides the industrial relevance, our work also contributes on a more technical level. Most studies derive uncertainty estimates for binary classification problems, we are only aware of a study of Roy et al. (2018) \cite{roy2018inherent}, which focuses on the task of deriving uncertainty measures in a multi-class image segmentation problem. Therefore, we contribute by deriving uncertainty measures for two further multi-class image segmentation problems. Additionally, we demonstrate, that a multiple linear regression can be applied to estimate segmentation quality in these multi-class segmentation problems. 
Regarding the challenge of estimating segmentation quality using uncertainty measures, we are only aware of DeVries \& Taylor (2018) \cite{devries2018leveraging}, who use a DNN to predict segmentation quality. Within our use case, we rely on a multiple linear regression model, as it is interpretable, and also can be used in scarce data settings. 
Additionally, researchers press for more insights on human-in-the-loop systems \cite{brynjolfsson2011race}, as successful designs are still scarce. Especially the allocation of (labeling) tasks between humans and machines is under-researched \cite{dellermann2019hybrid}. We contribute by implementing our human-in-the-loop system and evaluating it by a simulation study. To ensure generalizability, we assess the use of our approach on the publicly available Cityscapes \cite{cordts2016cityscapes} dataset for urban scene understanding.

\section{Foundations and Related Work}
First, we shortly introduce the motivation to use MC-dropout as an approach to estimate uncertainty. Subsequently, we present selected related studies, which focus on assessing the quality of a predicted segmentation by the use of uncertainty estimates.

There is a considerable body of literature growing around the theme of uncertainty estimation in DNNs. In classification tasks, a softmax output displays the probability of an output belonging to a particular class. Thus, softmax outputs are occasionally used to represent model uncertainty \cite{hendrycks2016baseline}. However, as illustrated by Gal (2015) \cite{gal2015dropout}, a model can be uncertain even with a high softmax output, indicating that softmax outputs do not represent model uncertainty accurately. Contrary to traditional machine learning approaches, a Bayesian perspective provides a more intuitive way of modeling uncertainty by generating a probability distribution over the outputs. However, inference in BNNs is challenging because the marginal probability can not be evaluated analytically, and, therefore, inference in BNNs is computationally intractable. Nevertheless, in a recent advance, Gal and Ghahramani (2015) \cite{gal2015dropout} show that taking Monte Carlo samples from a DNN in which dropout is applied at inference time approximates a BNN. In a study by Kendall \& Gal (2017) \cite{kendall2017uncertainties}, the authors show that these approximated BNNs lead to an improvement in uncertainty calibration compared to non-Bayesian approaches. Since dropout exists already in many architectures for regularization purposes, MC-dropout presents a scalable and straightforward way of performing approximated Bayesian Inference using DNNs without the need to change the training paradigm.

For the human-in-the-loop system, we are particularly interested in the prediction of segmentation quality based on uncertainty estimates. To date, several studies on this particular topic have been conducted within the field of medical imaging.
DeVries \& Taylor (2018) \cite{devries2018leveraging} use MC-dropout as a source of uncertainty to predict segmentation quality within the task of skin lesions segmentation. A separate DNN is trained to predict segmentation quality based on the original input image, the prediction output, and the uncertainty estimate. While the subsequent DNN achieves promising results in predicting segmentation quality, the subsequent model lacks transparency and explainability itself. In particular, the subsequent model does not provide any information why a prediction failed. Furthermore, a DNN is only applicable if a considerable amount of data is available.
Nair et al. (2020) \cite{nair2020exploring} utilize MC-dropout to explore the use of different uncertainty measures for Multiple Sclerosis lesion segmentation in 3D MRI sequences. The authors show that for small lesion detection, performance increases by filtering out regions of high uncertainty. 
While the majority of studies focus on pixel-wise uncertainties, there is a need to aggregate uncertainty on whole segments of an image. These aggregated structure-wise uncertainty measures allow an uncertainty assessment on an image-level.
The work of Roy et al. (2018) \cite{roy2018inherent} introduces three structure-wise uncertainty measures, also based on MC-dropout, for brain segmentation. While the authors show that these uncertainty measures correlate with prediction accuracy, the work does not display how these uncertainty measures are applicable in a broader context, e.g., in a human-in-the-loop system.

\section{Methodology}
On the basis of the previously presented foundations, we now introduce our applied methodology. In Subsection 3.1, we shortly introduce foundations regarding image segmentation. Subsequently, in Subsection 3.2, we present the modified U-Net architecture, which we use to approximate a BNN. Subsection 3.3 then describes the loss function and the evaluation metric, which we use to train and evaluate the modified U-Net. Lastly, Subsection 3.4 depicts the computation of the uncertainty measures on which our human-in-the-loop system relies.

\subsection{Image Segmentation} 
For our use case of wear analysis in the machining industry, information must be available on a pixel level to facilitate the assessment of location and size of wear for a given input image. An approach that provides this detailed information is called \textit{image segmentation}. While image segmentation can be performed unsupervised, it is often exercised as a supervised learning problem \cite{garcia2017review}. In supervised learning problems, the task is generally to learn a function $ f: X \rightarrow Y $ mapping some input values ($X$) to output values ($Y$). In the case of image segmentation, the concrete task is to approximate a function $f$, which takes an image $x$ as input and produces a segmentation $ \hat{y} $. The predicted segmentation assigns a category label $c \in C$ to each pixel $i \in N$ in the input image, where $C$ denotes the possible classes and $N$ the number of pixels in an input image. Therefore, the task of image segmentation is also referred to as pixel-wise classification. Consequently, the outputs must have the same height and width as the input image, the depth is defined by the number of possible classes.

\subsection{Model: Dropout U-Net} 
We apply a modified U-Net architecture due to its ability to produce good results even with a small amount of labeled images \cite{ronneberger2015u}. To avoid overfitting and increase performance, we implement the following adaptions to the original U-Net architecture: We use an L2-regularization in every convolutional layer, and additionally, we reduce the number of feature maps in the model's architecture, starting with 32 feature maps instead of 64 feature maps in the first layer \cite{bishop1995regularization}. The number of feature maps in the remaining layers follows the suggested approach from the original U-Net architecture, which doubles the number of feature maps in the contracting path and then analogically halves the number of feature maps in the expansive path. To accelerate the learning process, we add batch normalization between each pair of convolutional layers \cite{ioffe2015batch}. We use a softmax activation function in the last layer of the model to obtain predictions in the range [0,1]. Therefore, the modified U-Net takes an input image $x$ and produces softmax probabilities $p_{i,c} \in \left [0,1  \right ]$ for each pixel $i \in N$ and class $c \in C$, compare equation (1).
\begin{equation}
p_{i,c} = f(x) \quad \quad \forall i \in N, \forall c \in C
    \label{eq:softmax_outputs}
\end{equation}
As a source of uncertainty, we realize the human-in-the-loop system based on MC-dropout because of its implementation simplicity, while still being able to generate reasonable uncertainty estimates \cite{gal2015dropout}. We employ dropout layers in the modified U-Net to enhance the model with the ability to approximate a BNN \cite{gal2015dropout}. Units within the dropout layers have a probability of 0.5 to be multiplied with zero and therefore, to drop out. We follow the suggested approach by Kendall et al. (2015) \cite{kendall2015bayesian_segnet} in the context of the Bayesian SegNet to use dropout layers at the five most inner decoder-encoder blocks. Hereinafter, we will refer to the modified U-Net, which applies dropout at inference time, as the \textit{Dropout U-Net}. By the application of dropout at inference time, the Dropout U-Net constitutes a stochastic function $f$. For multiple stochastic forward passes $T$ of an input image $x$, the Dropout U-Net generates a probability distribution for each $p_{i,c}$. To obtain a segmentation $\hat{y} $, we calculate the mean softmax probability first, as described in equation (\ref{eq:mean_dropout}). Then, the segmentation $\hat{y}$ is derived by applying the argmax function over the possible classes of the softmax probabilities $p_{i,c}$, compare equation (\ref{eq:softmax_to_pred}).

\begin{align}
    \label{eq:mean_dropout}
         p_{i,c} = \frac{1}{T} \sum_{t=1}^{T} p_{i,c,t} \quad \quad \forall i \in N, \forall c \in C 
\end{align}
\begin{equation}
\hat{y_i} = \underset{c \in C}{\operatorname{argmax}}(p_{i, c})  \quad \quad \forall i \in N
    \label{eq:softmax_to_pred}
\end{equation}

\subsection{Loss Function and Performance Evaluation Metric}
As a loss function, we use a \textit{weighted category cross-entropy} loss. We weight each class with the inverse of its occurrence (pixels) in the training data due to class imbalance. This leads to an equal weighting between the classes in the loss function \cite{crum2006generalized}. The weighted categorical cross-entropy loss is defined in equation (\ref{eq:loss_function}); $g_{i,c}$ denotes the one-hot-encoded ground truth label, and $w_c$ the computed weights for each class.
\begin{equation}
\label{eq:loss_function}
\mathcal{L} (p, g) = - \frac{1}{N} \sum_{i=1}^{N} {\sum_{c=1}^{C}{w_{c} \;  g_{ i,c } \; \log \; p_{i,c}  }}
\end{equation}
To reflect the quality of a predicted segmentation, we rely on the \textit{Dice coefficient}  \cite{dice1945measures} (DSC) as a performance evaluation metric. The Dice coefficient assesses the overlap, or intersection, between the model's outputs and the one-hot-encoded ground truth labels $g_{i,c}$. A full overlap between a prediction and a label is represented by a value of one. If there is no overlap, the Dice coefficient returns zero. Representing the outputs, one could use the softmax probabilities $p_{i,c}$ or the binarized one-hot encoded predictions $\hat{y}_{i,c} $. We use the binarized predictions $\hat{y}_{i,c} $ since these predictions represent the foundation for the build-on tool wear analysis.  
As suggested by Garcia et al. (2017) \cite{garcia2017review} for the related Jacard-Coefficient, we compute the Dice coefficient for each class separately. To assess the segmentation quality of an input image, we compute the averaged Dice coefficient across all classes, leading to a \textit{mean Dice coefficient}, defined in equation (\ref{eq:averaged_dice}). To evaluate a model on the test set, we calculate the average across the mean Dice coefficients per image. Next to the Dice coefficient, we also compute the pixel accuracy as a performance measure. It defines the percentage of correctly classified pixels.
\begin{equation}
\mathit{Mean \; Dice  \; Coefficient} = \frac{2}{C} \sum_{c=1}^{C} \frac{\sum_{i=1}^{N} \hat{y}_{i,c} \; g_{i,c} }{\sum_{i=1}^{N} \hat{y}_{i,c}  + \sum_{i=1}^{N} g_{i,c}}
    \label{eq:averaged_dice}
\end{equation}
\subsection{Uncertainty Estimation} \label{Methodology: Uncertainty Estimation}

As a next step, we describe how uncertainty measures can be calculated using the probability distribution outputs of the Dropout U-Net. In general, the information-theory concept of entropy \cite{shannon1948mathematical} displays the expected amount of information contained in the possible realizations of a probability distribution. Following previous work \cite{kendall2017uncertainties}, we utilize the entropy as an uncertainty measure to reflect the uncertainty of each pixel in a predicted segmentation:
\begin{equation}
    H(p_i) = -  \sum_{c=1}^{C} \; p_{i, c} \; \log \; p_{i, c} \quad \quad \forall i \in N 
    \label{eq:unc_pred_entropy}
\end{equation}
The entropy displays its maximum if all classes have equal softmax probability and it reaches its minimum of zero if one class holds a probability of $1$ while the other classes have a probability of $0$. Therefore, the entropy reflects the uncertainty of a final output $\hat{y_i}$ by considering the model's outputs $p_{i,c}$ over all classes $C$. For the task of image segmentation, the entropy is available per pixel. However, for several applications, it is necessary to derive an uncertainty estimate on a higher aggregation level. For example, within the tool wear analysis, we want to decide on an image basis, whether a segmentation is successful or failed. One approach is to average the pixel-wise entropy values over an image to come up with an image-wise uncertainty estimate. Roy et al. (2018) \cite{roy2018inherent} propose to calculate the average pixel uncertainty for each predicted class in a segmentation. We utilize this idea in the context of wear analysis and define the entropy per predicted class  $U_c$ in equation (\ref{eq:unc_pred_entropy_1}). Equation (\ref{eq:unc_pred_entropy_2}) defines the number of pixels for each predicted class. The entropy per predicted class provides information on an aggregated class-level and can be used to estimate uncertainty for each class in an input image. Since only the predictions are used, this uncertainty estimate is applicable in the absence of ground truth.
\begin{align}
    U_c = \frac{1}{S_c} \sum_{i=1}^{S_c} H(p_i) \quad \quad \forall c \in C
        \label{eq:unc_pred_entropy_1}
    \\
    S_c = \left \{i \in N \; | \; \hat{y_i} = c \right \} \quad \quad \forall c \in C
    \label{eq:unc_pred_entropy_2}
\end{align}

\section{Experiments}
With the methodology at hand, the upcoming Subsection 4.1 provides information about the two datasets, the preprocessing and training procedures. Subsequently, Subsection 4.2 briefly describes the performance results in terms of the Dice coefficient. In Subsection 4.3, we evaluate the uncertainty-based human-in-the-loop system in the following way: 
First, we illustrate the relation between uncertainty and segmentation quality using two exemplary predictions. Then, we quantitatively assess the relation between uncertainty and segmentation quality using the Bravais-Pearson correlation coefficient. Next, we fit a multiple linear regression on each test set, which uses the uncertainty measures as independent variables to explain segmentation quality. Lastly, we simulate the performance of the uncertainty-based human-in-the-loop system based on the multiple regression and compare it against a random-based human-in-the-loop system.

\subsection{Datasets, Preprocessing and Training Procedure}
The unique \textbf{Tool wear dataset} consists of 213 pixel-wise annotated images of cutting tool inserts, which were previously used by Ceratizit's customers in real manufacturing processes until their end-of-life. The labels are created as follows: The first 20 images are labeled jointly by two domain experts from Ceratizit. Afterward, labels are assigned individually, whereas at least two domain experts discuss unclear cases. The recording of the images is standardized to reduce the required amount of generalization of the learning algorithm. The images initially have a resolution of 1600$\times$1200 pixels. As a preprocessing step, we cut the image to a shape of 1600$\times$300, which lets us focus on the cutting edge where the wear occurs. For computational efficiency and as a requirement of the U-Net architecture, we resize the images to a shape of 1280$\times$160 using a bilinear interpolation. We randomly split the dataset into 152 training, 10 validation and 51 test images. 
The model is trained for 200 epochs, using an Adam optimizer, a learning rate of 0.00001, an L2-regularization with an alpha of 0.01, and a batch size of 1. We choose the hyperparameters after running a brief hyperparameter search. The training is conducted on a Tesla V100-SXM2 GPU, with a training duration of approximately two hours. In the literature, the number of conducted forward passes ranges from 10 to 100 \cite{devries2018leveraging}, we use 30 Monte Carlo forward passes to create the probability distribution over the outputs.

The \textbf{Cityscapes dataset}  \cite{cordts2016cityscapes} is a large-scale dataset, which contains images of urban street scenes from 50 different cities. It can be used to assess the performance of vision-based approaches for urban scene understanding on a pixel-level. The Cityscapes dataset initially consists of 3475 pixel-wise annotated images for training and validation. Performance is usually specified through 1525 test set images for which ground truth labels are only available on the Cityscapes website \cite{cordts2016cityscapes}. Since we need ground truth labels to assess the uncertainty estimates, we use the 500 proposed validation images as the test set and randomly split the remaining 2975 images into 2725 training and 250 validation images. There are initially 30 different classes which belong to eight categories. We group classes belonging to the same category together, to create a similar problem setting between the Cityscapes dataset and the Tool wear dataset. 
Furthermore, we combine the categories 'void',  'object', 'human', and 'nature' to one class, which we consider in the following as the background class. The remaining categories are 'flat', 'construction', 'sky' and 'vehicle', ultimately resulting in five different classes. 
The original images have a resolution of 2048$\times$1024 pixels. During training, an augmentation step flips the images horizontally with a probability of 0.5. Then, a subsequent computation randomly crops the images with a probability of 0.15 to an input size of 1024$\times$512. Otherwise, the images are resized using a bilinear interpolation to the desired input shape of 1024x512. Following a brief hyperparameter search, we train the model for 15 epochs with a learning rate of 0.0001 using an Adam optimizer, a batch size of 10, and an L2 weight regularization with an alpha of 0.02. We train the model on a Tesla V100-SXM2 GPU for approximately 2.5 hours. The Dropout U-Net uses five Monte Carlo forward passes, considering the higher computational complexity due to the more extensive test set and the large image size.

\subsection{Performance Results}
Next, we assess the quality of a predicted segmentation in terms of the Dice coefficient. Table \ref{performance_results} displays the performance results on each respective test set.
The background class has the largest proportion of pixels (93.9\%) in the tool wear dataset, followed by flank wear (4.7\%), built-up edge (0.9\%) and lastly, chipping (0.4\%). We assume that the amount of labeled pixels of a specific class is closely related to prediction performance. We find, that the model has particular difficulties segmenting chipping phenomena, which is expressed by a Dice coefficient of 0.244. We explain this lack of prediction performance by the punctual and minor occurrence of chipping phenomena within images, compare Figure \ref{fig_seg_map}, and the characteristic of the Dice coefficient. In particular, the Dice coefficient per class drops to zero, if the model produces a false negative, meaning the model falsely predicts a wear mechanism, and if there is no wear for the corresponding class labeled in the image. This characteristic and the challenging task of classifying small chipping phenomena in an input image causes the Dice coefficient of the chipping class to drop to zero for several images. In contrast to chipping, the Dropout U-Net recognizes the background class well and is classifying flank wear and built-up edge considerably well. Compare the label (\ref{fig_seg_map_c}) and the prediction (\ref{fig_seg_map_e}) for an illustration. As illustrated on the left hand side of Figure \ref{fig_seg_map}, the input image is predicted well as the intersection between label and prediction is high. The performance results of the Cityscapes dataset also indicate that the Dropout U-Net can generate a predicted segmentation for each class considerably well. 

\begin{table}[]
    \caption{Performance results}
    \label{performance_results}
    \begin{minipage}{.5\linewidth}
      \centering
        \begin{tabular}{lc}
        \multicolumn{2}{c}{Tool wear dataset} \\ 
        \toprule
            Class & \thead{Dice coefficient} \\
            \midrule
            Background  	& 	0.991\\
            Flank wear 		  &   0.695	\\
            Chipping 		   &   0.244	\\
            Built-up edge   &    0.596 \\
            &  \\
            \midrule
            \midrule
            Mean DSC &  0.631 \\
            Pixel accuracy 		  & 0.977 \\ 
            \bottomrule
        \end{tabular}
    \end{minipage}%
    \begin{minipage}{.5\linewidth}
      \centering
        \begin{tabular}{lc}
        \multicolumn{2}{c}{Cityscapes dataset} \\ 
        \toprule
            Class & \thead{Dice coefficient} \\
            \midrule
            Background						&  0.929 \\
            Flat 										 & 0.693 \\
            Construction						& 0.830 \\
            Sky 									  & 0.769 \\
            Vehicle 							   & 0.773 \\
            \midrule
            \midrule
            Mean DSC 								& 0.799 \\
            Pixel accuracy 							& 0.875 \\ 
            \bottomrule
        \end{tabular}
    \end{minipage} 
\end{table}

\subsection{Evaluation: Uncertainty-based Human-in-the-loop System}
\begin{figure}[ht] 
      \label{performance_results_tool_wear}
  \begin{subfigure}[b]{0.5\linewidth}
    \centering
    \includegraphics[width=0.95\linewidth]{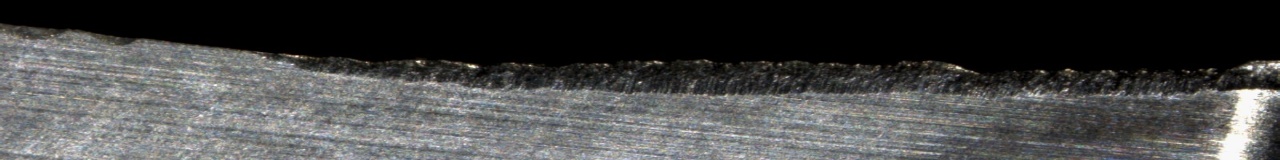} 
    \caption{Preprocessed input image (1)} 
    \label{fig_seg_map_a} 
    \vspace{2ex}
  \end{subfigure}
  \begin{subfigure}[b]{0.5\linewidth}
    \centering
    \includegraphics[width=0.95\linewidth]{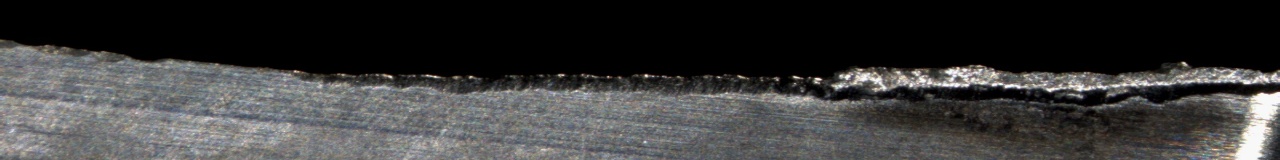} 
    \caption{Preprocessed input image (2)} 
    \label{fig_seg_map_b} 
    \vspace{2ex}
  \end{subfigure} 
  \begin{subfigure}[b]{0.5\linewidth}
    \centering
    \includegraphics[width=0.95\linewidth]{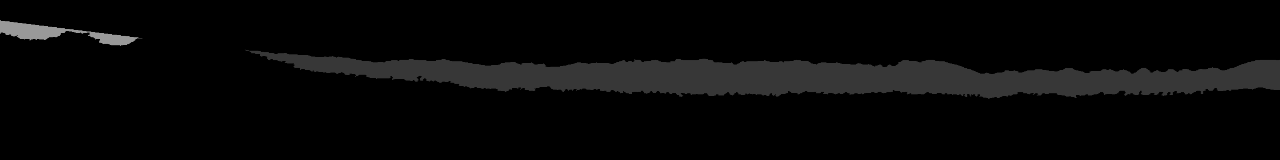} 
    \caption{Label (1)} 
    \label{fig_seg_map_c}
    \vspace{2ex}
  \end{subfigure}
  \begin{subfigure}[b]{0.5\linewidth}
    \centering
    \includegraphics[width=0.95\linewidth]{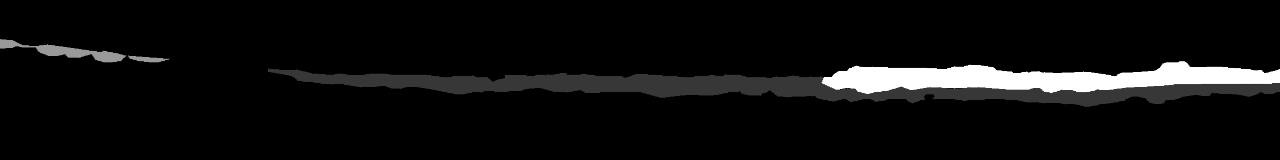} 
    \caption{Label (2)} 
    \label{fig_seg_mamp_d} 
    \vspace{2ex}
  \end{subfigure} 
  \begin{subfigure}[b]{0.5\linewidth}
    \centering
    \includegraphics[width=0.95\linewidth]{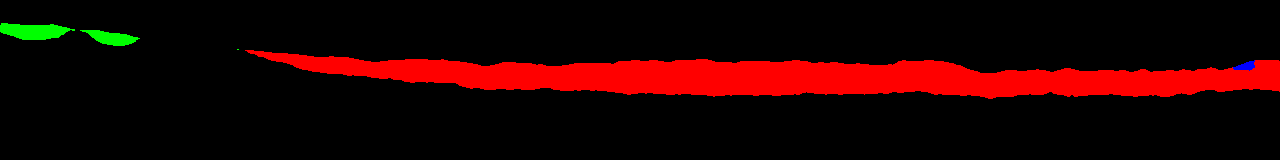} 
    \caption{Prediction (1)} 
    \label{fig_seg_map_e} 
    \vspace{2ex}
  \end{subfigure}
  \begin{subfigure}[b]{0.5\linewidth}
    \centering
    \includegraphics[width=0.95\linewidth]{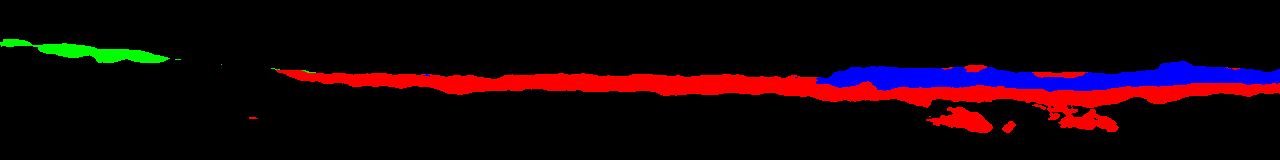} 
    \caption{Prediction (2)} 
    \label{fig_seg_map_f} 
    \vspace{2ex}
  \end{subfigure} 
  \begin{subfigure}[b]{0.5\linewidth}
    \centering
    \includegraphics[width=0.95\linewidth]{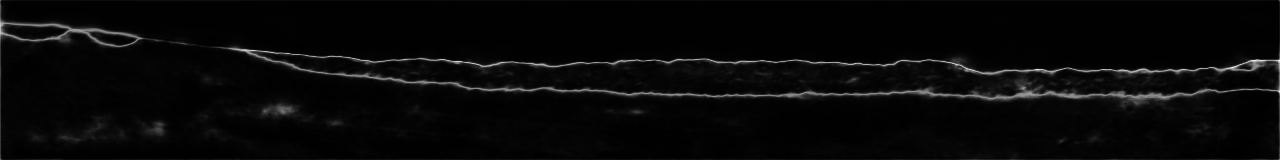} 
    \caption{Entropy per pixel (1)} 
    \label{fig_seg_map_g} 
  \end{subfigure}
  \begin{subfigure}[b]{0.5\linewidth}
    \centering
    \includegraphics[width=0.95\linewidth]{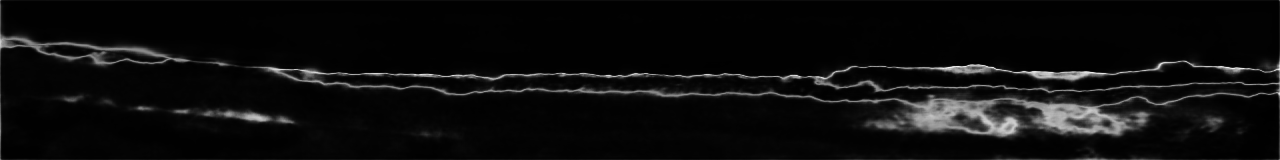} 
    \caption{Entropy per pixel (2)} 
    \label{fig_seg_map_h} 
  \end{subfigure} 
  \caption{Illustration of two images of the test set with their corresponding labels, predictions and uncertainty maps (best viewed in color).\\ Color coding: Flank wear = dark grey/red, Chipping = light grey/green and Built-up edge = white/blue}
  \label{fig_seg_map} 
\end{figure}

With the performance results at hand, we focus on assessing the uncertainty-based human-in-the-loop system. Figure \ref{fig_seg_map} presents two preprocessed images, their corresponding human labels, their predictions, and the generated uncertainty maps for the Tool wear dataset. The uncertainty maps are generated using pixel-wise uncertainties based on the entropy, compare equation (\ref{eq:unc_pred_entropy}). Within the uncertainty maps, brighter pixels represent uncertain outputs, and darker pixels represent certain predictions. The uncertainty map (g) of the input image (1) displays uncertain outputs, indicated by brighter pixels, at the edge between classes. This behavior is often noticed within uncertainty observation in image segmentation tasks \cite{kendall2015bayesian_segnet}, as it reflects the ambiguity of defining precise class regions on a pixel-level. The most interesting aspect of Figure \ref{fig_seg_map} is the prediction (f) and the corresponding uncertainty map (h). The model falsely predicted several areas on the right hand side as flank wear (red). However, the model also indicates high uncertainty for this particular area, indicated by brighter pixels in the corresponding uncertainty map (h).

This relationship is essentially the foundation of the uncertainty-based human-in-the-loop system. The relationship between a pixel's uncertainty and the probability of being classified correctly enables the system to distinguish between images which the model segments successfully and images which the model segments poorly. As the goal is to distinguish segmentation quality on an image level, the average uncertainty of a predicted segmentation can be used as an uncertainty measure for an image. However, we find that for the task of tool wear, there does not exist a significant correlation ($-$0.34) between the averaged entropy per prediction and the mean Dice coefficient on the 51 images of the test set. 
The same analysis yields a correlation of $-$0.57 for the Cityscapes dataset. However, we find that the mean entropy of a predicted class, $U_c$, is highly correlated with the corresponding Dice coefficient per class. Table \ref{correlation_results} displays the Bravais-Pearson Correlation between the entropy of a predicted class and the corresponding Dice coefficient per class on the respective test sets. In the case of the Tool wear dataset, the linear relationship is especially strong for the classes flank wear, chipping and built-up edge, while it is slightly weaker for the background class. These results are reproducible on the Cityscapes dataset. As can be seen from Table \ref{correlation_results}, correlations, besides the background class, range from $-$0.751 to $-$0.878, indicating a strong negative linear relationship.
\begin{table}[]
    \caption{Correlation coefficients between the uncertainty per predicted class ($U_c$) and the Dice coefficients per class}
    \label{correlation_results}
    \begin{minipage}{.5\linewidth}
      \centering
        \begin{tabular}{lc}
        \hfill \\
        \multicolumn{2}{c}{Tool wear dataset } \\
            \toprule
            Class & Correlation \\
            \midrule
            Background 		  & -0.656 \\
            Flank wear 			& -0.911	\\
            Chipping 			 & -0.818 \\
            Built-up edge 	 & -0.932\\
            & \\
            \bottomrule
        \end{tabular}
    \end{minipage}%
    \begin{minipage}{.5\linewidth}
        \centering
            \begin{tabular}{lc}
            \hfill \\
            \multicolumn{2}{c}{Cityscapes dataset} \\ 
            \toprule
            Class & Correlation \\
            \midrule
             Background 	& -0.208 \\
            Flat					 & -0.878	\\
            Construction  & -0.754 \\
            Sky				  & -0.751 \\
            Vehicle 			& -0.858 \\
            \bottomrule
        \end{tabular}
    \end{minipage} 
\end{table}
We use ordinary least squares to fit a multiple linear regression on the test set for both datasets using the uncertainties per predicted class $U_c$ as independent variables and the mean Dice coefficient as the dependent variable. Subsequently, the linear regression aims at quantifying the prediction quality on an image-level in the absence of ground truth. We find that, for both datasets, the independent variable 'uncertainty per background class' is statistically not significant at the 0.05 value. Therefore, we discard it as an independent variable from the multiple regression model. The remaining independent variables are significant at the 0.01 level for both datasets. The regression results yield a $R^2=0.718$ for the Tool wear dataset and a $R^2=0.655$ for the Cityscapes dataset. This indicates that the multiple linear regression can explain a substantial amount of variation of the mean Dice coefficient. The full regression results can be found in the appendix.

\begin{figure}[ht] 
      \label{performance_results_tool_wear}
  \begin{subfigure}[b]{0.5\linewidth}
    \centering
    \includegraphics[width=0.95\linewidth]{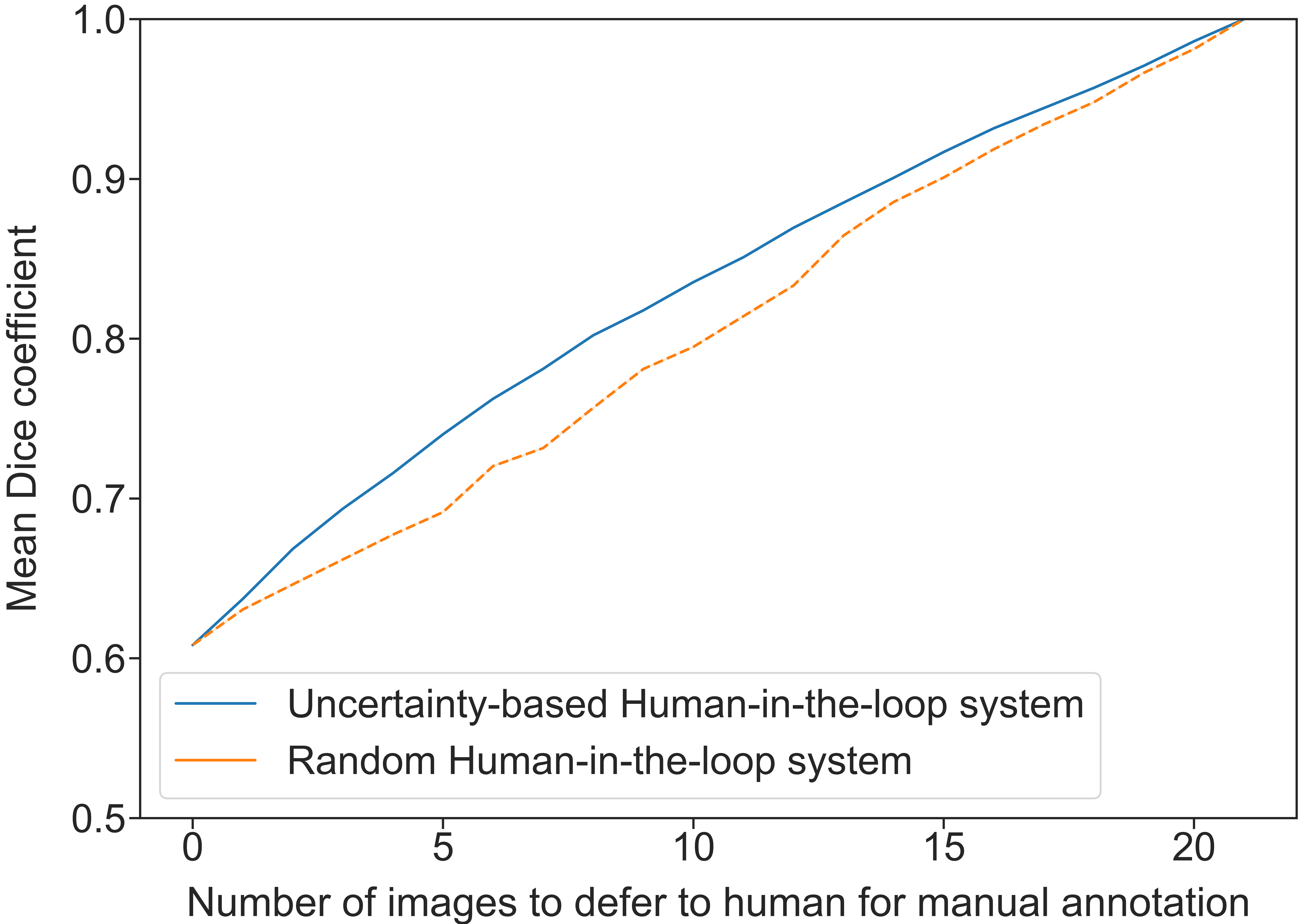}
    \caption{Tool wear dataset} 
    \label{fig_simulation_tool_wear} 
  \end{subfigure}
  \begin{subfigure}[b]{0.5\linewidth}
    \centering
    \includegraphics[width=0.95\linewidth]{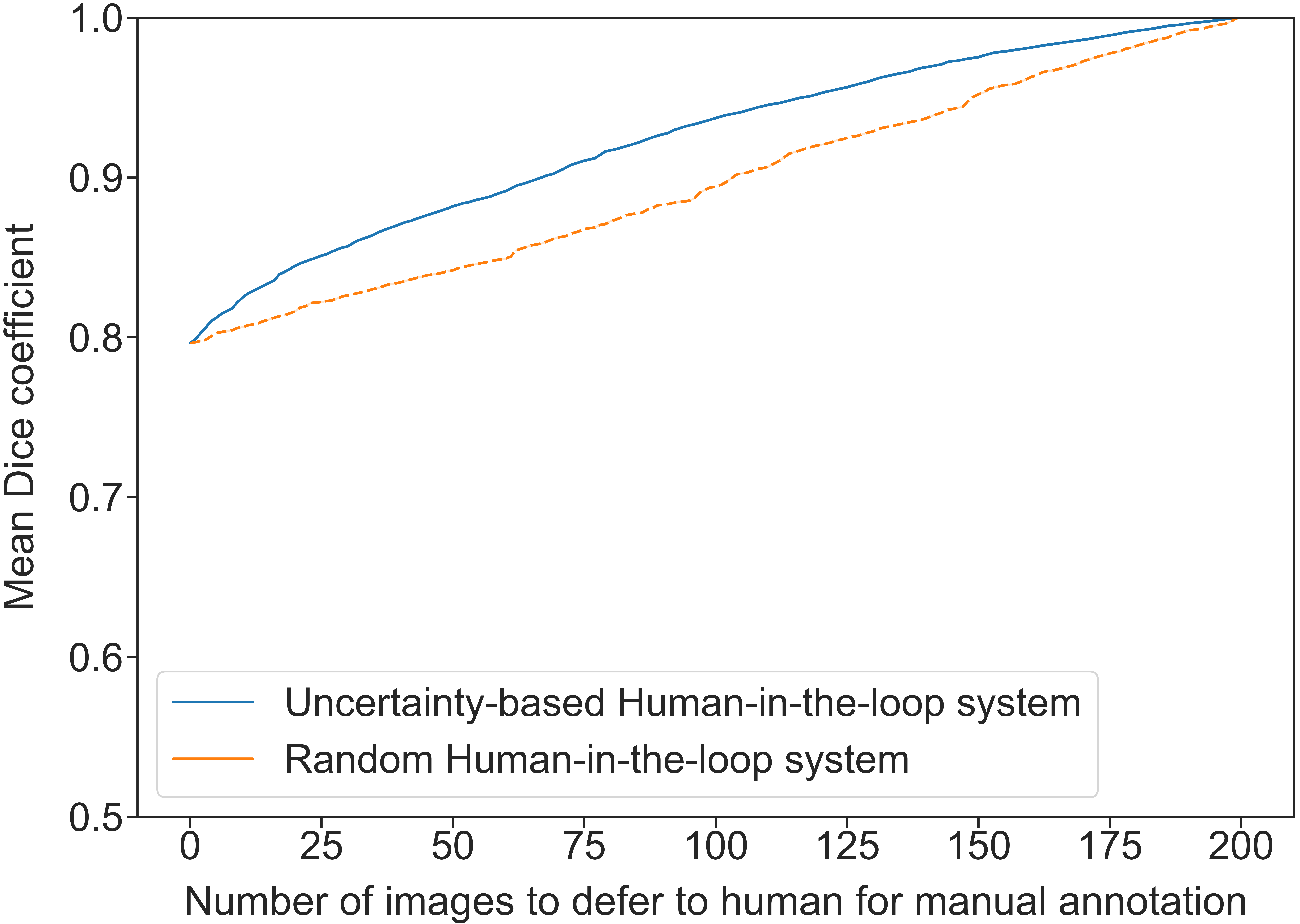} 
    \caption{Cityscapes dataset} 
    \label{fig_simulation_cityscapes} 
  \end{subfigure}
    \caption{Simulation results}
  \label{fig_simulation_results} 
\end{figure}

In the following paragraph, we assess the use of the multiple linear regression in the context of a human-in-the-loop system by running a simulation. The multiple linear regression predicts the quality of a predicted segmentation in terms of the mean Dice coefficient, using the uncertainties per predicted class as independent variables. Then, in an iterative process, the input image, for which the prediction displays the lowest estimated mean Dice coefficient is forwarded for human annotation. Images, displaying a higher estimated Dice coefficient are retained by the system. Within the simulation, we assume a perfect human segmentation, and set the corresponding Dice coefficient of the forwarded image to one. The performance of the system is then calculated by combining the mean Dice coefficients of the retained images and the forwarded human annotated images. Figure \ref{fig_simulation_results} shows the simulation results for both datasets. The x-axis displays the number of images, which are forwarded to human annotation. The y-axis displays the performance of the system. We compare the performance of the human-in-the-loop system (blue line) against a random-based human-in-the-loop sytem (orange line). Contrary to the uncertainty-based system, the random-based system decides randomly, which images are forwarded to human annotation. To avoid overfitting, we use a split for each dataset as follows: For the Tool wear dataset, the multiple linear regression is fitted on 30 images, the remaining 21 predictions are then used for simulation. Within the Cityscapes dataset, we use 300 test set images to fit the regression and 200 images for simulation. For both datasets, the uncertainty-based human-in-the-loop system is able to achieve a better mean Dice coefficient using less human annotations than a random-based approach. This is due to the multiple regression, which identifies low-quality predictions and therefore enables the system to forward these predictions to human annotation first.  

\section{Discussion and Outlook}
In this work, we show the applicability and usefulness of an uncertainty-based human-in-the-loop system for the task of industrial tool wear analysis. The human-in-the-loop system addresses critical challenges regarding the adoption of CNNs in industry.
In particular, it increases transparency by providing uncertainty measures which are correlated with segmentation performance. Additionally, it improves performance by incorporating a human expert for the annotation of images that are estimated to be of low quality.

Within the use case of tool wear analysis, we consider the task of segmenting three different wear mechanisms on cutting tool inserts. We apply and train a modified U-Net architecture on a real-world dataset of our industry partner Ceratizit, achieving good performance results.  For the human-in-the-loop system, we enhance the existing tool wear analysis with the following capabilities: We implement MC-dropout and use the information-theory concept of entropy to compute pixel-wise uncertainties. Furthermore, we aggregate the pixel-wise uncertainties to compute class-wise uncertainty measures on an image-level. A multiple linear regression reveals that the class-wise uncertainties can be used as independent variables to explain a substantial amount of the mean Dice coefficient of an image. The multiple linear regression is then leveraged within the human-in-the-loop system to decide, whether a given segmentation should be forwarded to a human expert, or be retained in the system as a successful prediction. A simulation study demonstrates that the performance improves through the utilization of a human expert, which annotates estimated low-quality predictions. Furthermore, the system increases transparency by additionally issuing an estimate about the quality of a prediction. 

We assess our system not only on our proprietary tool wear data set but also on the publicly available and substantially larger Cityscapes data set, confirming the generalizability of our approach to the task of urban scene understanding. Nevertheless, we consider the application to only two data sets as a limitation of the study. In the future, we aim at validating the system on additional datasets.
Another promising avenue for future research is to further distinguish uncertainty into epistemic (model) and aleatoric (data) uncertainty \cite{kendall2017uncertainties}. While aleatoric uncertainty is due to inherent noise in the input data, e.g., a blurred image, epistemic uncertainty occurs due to model uncertainty, e.g., lack of training data. Within human-in-the-loop systems, this distinction can lead to more informed decision, e.g., when images are forwarded to a human expert, a possible cause for a failed prediction can be provided. 
Regarding uncertainty estimation, further research is also needed on a more theoretical level, to establish a more profound understanding of uncertainty outputs of different approaches and their relation to prediction quality.  This could include a structured comparison of different ways to calculate uncertainty across different use cases. 
Lastly, from a human-centric machine learning standpoint, further research should assess, if the increased transparency of the human-in-the-loop system leads to a more calibrated level of trust from the user. While we found several indications in the literature, few studies have investigated this relationship in a systematic way.

We see a broad applicability of the uncertainty-based human-in-the-loop system in industrial applications. While we consider the task of image segmentation, the general observations should be relevant in a variety of supervised learning problems.
A human-in-the-loop system can be beneficial for all types of automation tasks, in which human experts display superior performance than automated systems, but in which the automated system is more cost efficient. An example for such a system would be an industrial quality control system. Otherwise, we perceive limited potential for tasks, in which the performance of human experts is inferior compared to the performance of automated systems. This scenario would include many applications of time series forecasting. In these tasks, the estimated prediction quality could only be used to issue warnings whenever an output is likely to be faulty.
Altogether, we believe that the uncertainty-based human-in-the-loop system represents an essential building block for the facilitation of a more widespread adoption of CNN-based systems in the industry.
\section*{Acknowledgments} We would like to thank Ceratizit Austria GmbH, in particular Adrian Weber for facilitating and supporting this research.

%

\section{Appendix}
\begin{table}[]
    \caption{Regression results}
    \begin{minipage}{.5\linewidth}
      \label{regression_results}
      \centering
        \begin{tabular}{lc}
            \multicolumn{2}{c}{Tool wear dataset}\\
            \toprule
             \multicolumn{2}{c}{\textit{Dependent variable: Averaged Dice}} 		\tabularnewline
			\midrule
			Const & 0.922$^{***}$  \tabularnewline
			Background &   \tabularnewline
			Flank wear &  -0.165$^{***}$  \tabularnewline
			Chipping &  -0.099$^{***}$  \tabularnewline
			Built-up edge & -0.169$^{***}$   \tabularnewline
            & \tabularnewline
			\midrule
			Observations        & 51  \tabularnewline
			R${2}$              & 0.718  \tabularnewline
			Adjusted R${2}$     & 0.7  \tabularnewline
			Residual Std. Error & 0.066   \tabularnewline
			F Statistic         & 39.849$^{***}$  \tabularnewline
		\bottomrule \tabularnewline
        \end{tabular}
    \end{minipage}%
    \begin{minipage}{.5\linewidth}
        \centering
            \begin{tabular}{lc}
            \multicolumn{2}{c}{Cityscapes dataset} \\
            \toprule
             \multicolumn{2}{c}{\textit{Dependent variable: Averaged Dice}} \tabularnewline
			\midrule
			Const  &  1.206$^{***}$  \tabularnewline
			Background & \tabularnewline
            Flat & -0.172$^{***}$	\tabularnewline
			Construction & -0.203$^{***}$ \tabularnewline
			Sky & -0.131$^{***}$ \tabularnewline
			Vehicle & -0.132$^{***}$ \tabularnewline
			\midrule
			Observations         & 500 \tabularnewline
             R${2}$              			& 0.655 \tabularnewline
			Adjusted R${2}$     	 & 0.653 \tabularnewline
			Residual Std. Error  & 0.078  \tabularnewline
			F Statistic           & 235.335$^{***}$ \tabularnewline
		\bottomrule \tabularnewline
        \end{tabular}
    \end{minipage}
    \centerline{Note: $^{*}\, p<0.1$; $^{**}\, p<0.05$; $^{***}\, p<0.01$}
\end{table}


\end{document}